# How I won the "Chess Ratings — Elo vs the Rest of the World" Competition

Yannis Sismanis

December 2010

## Abstract

This article discusses in detail the rating system that won the kaggle competition "Chess Ratings: Elo vs the rest of the world". The competition provided a historical dataset of outcomes for chess games, and aimed to discover whether novel approaches can predict the outcomes of future games, more accurately than the well-known Elo rating system. The winning rating system, called Elo++ in the rest of the article, builds upon the Elo rating system. Like Elo, Elo++ uses a single rating per player and predicts the outcome of a game, by using a logistic curve over the difference in ratings of the players. The major component of Elo++ is a regularization technique that avoids overfitting these ratings. The dataset of chess games and outcomes is relatively small and one has to be careful not to draw "too many conclusions" out of the limited data. Many approaches tested in the competition showed signs of such an overfitting. The leader-board was dominated by attempts that did a very good job on a small test dataset, but couldn't generalize well on the private hold-out dataset. The Elo++ regularization takes into account the number of games per player, the recency of these games and the ratings of the opponents. Finally, Elo++ employs a stochastic gradient descent scheme for training the ratings, and uses only two global parameters (white's advantage and regularization constant) that are optimized using cross-validation.

## 1 Introduction

The kaggle dataset [5] consists of more than 73 thousand month-stamped game outcomes between roughly 8 thousand distinct players. The competition was designed in the following way: A hold-out dataset of about 8 thousand games was created using the last five months of the dataset. The remaining data consist the training dataset. Only for the training data the outcomes for the games were made known. The outcomes of the hold-out games, the identities of the players and their actual Elo [3] ratings, were kept secret until the end of the competition. A 20% sample of the hold-out dataset was used as a test dataset. The rest of the hold-out dataset was used as a private dataset. A competitor submits predictions (i.e. win, draw, loss) for all games in the hold-out set. The kaggle system computes a prediction performance metric, called the Player/Month-aggregated Root Mean Square Error (PM-RMSE), of the predictions over both the private and the test dataset. The test dataset PM-RMSE is returned to the competitor as an indication of the submission's performance and is posted on a leader-board. Finally, the winners of the competition are the ones that performed best on the private dataset. A similar technique was used during the Netflix competition [6] to preclude "clever" systems from probing repeatedly the hold-out dataset and thus "gaming" the competition.

Chess Ratings systems (a nice introduction is in [1]) learn the ratings of each player, by fitting the observed outcomes of the games in the training dataset. The goal is to generalize the ratings in a way that best allows the prediction of future unknown outcomes (those in the hold-out dataset). Since the given datasets are relatively small, extreme caution is required to avoid overfitting the ratings over both the training dataset and the test dataset. During the competition, the leader-board was dominated by approaches that performed extremely well on the test dataset but didn't generalize well on the private hold-out dataset. In this article, we present the winning rating system inspired by Elo [3] and Chessmetrics [2]. We call the winning system Elo++, and it's main extension is that it employs an $l_2$ regularization technique to avoid overfitting the ratings. The regularization takes into account the number of games per player, the recency of these games and the ratings of the opponents of each player. The intuition is that any rating system should "trust" more the ratings of players who have played a lot of recent games versus the ratings of players who have played a few old games. The extent of regularization in Elo++ is controlled using a single parameter, that is optimized through cross-validation.

One could try associating two ratings per player, depending on whether this player is playing white or black, or even use a large vector of ratings per player. However, introducing more ratings per player, multiplies the potential for overfitting. Judging by the outcome of the competition, overfitting is a big problem for rating systems. One should handle it correctly, even when using models with just a single rating per player. Elo++ is designed to use a single rating; it facilitates rating in general, is much more intuitive and the potential for enhancing existing Elo lists is greater.

Elo++ treats older games differently than newer games. Intuitively older games should affect less the current rating of a player than newer games. Various temporal dynamics affect the ratings of the players. For example, younger players get better much faster than others, established strong players demonstrate small fluctuations in their performance, and older players get progressively worse as time passes by. Capturing these temporal dynamics is important for any rating system; in Elo++ a simple weighting scheme over the recency of the games has been employed.

Most games between chess players happen in tournaments, where players of comparable strength play against each other. It is very rare, that a weak player participates in a tournament of strong players (and vice-versa). This observation implies that the rating of a player and the ratings of his opponents are strongly correlated. In Elo++ a weighted average rating of the opponents of each player is taken into account while training a players individual rating. Similar ideas have been proposed by Jeff Sonas —who put together this competition— in the Chessmetrics [2] rating system.

The complete Elo++ rating system was implemented in about 100 lines of R [9] code and employs a stochastic gradient descent technique for training the ratings. Elo++ made the best predictions in the competition, outperforming all approaches tested including benchmarks and variants of TrueSkill [7], Chessmetrics [2], Glicko [4], Elo [3] and others.

In the following sections, we discuss in detail the winning Elo++ rating system. First we introduce basic notation and terminology in Section 2. Then we discuss in detail the main ideas around Time Scaling(Section 3.1) and Neighbors(Section 3.2). Training, regularization and the exact iterative update rules for the ratings are discussed in Section 3.3. Elo++uses only two global parameters $\gamma$ (white's advantage) and $\lambda$ (regularization constant); their nature and optimal values are discussed in Section 3.4. In Section 4 we discuss the quality of the computed Elo++ ratings and compare against the corresponding Elo ratings.

## 2  Basics & Notation

We introduce a simple notation, for ease of presentation of the rating system. Each player $i$ is associated with a single rating $r_i$. The outcome of a game between the player $i$ with white, and player $j$ with black is denoted with $o_{ij}$. The outcome is 1 if white wins, 1/2 if it is a draw, and 0 if black wins. We distinguish between predicted outcomes from known ones, using the $\hat{o}_{ij}$ notation for the predicted ones.

The month a game happens between players $i$ and $j$ *regardless of color* is denoted as $t_{ij}$. Note, that two players may have played many games at different (or even the same) months. For ease of exposition, we'll use the simple notations $o_{ij}$, $\hat{o}_{ij}$ and $t_{ij}$ to refer to all such games. The training dataset $T$ consists of tuples of the form $\langle i, j, t_{ij}, o_{ij} \rangle$. A submission consists of predictions of the form $\langle i, j, t_{ij}, \hat{o}_{ij} \rangle$, for all the games in the hold-out dataset. The actual game outcomes $o_{ij}$ for the hold-out dataset were kept secret until the end of the competition. We denote as $D$ the domain of $i$ and $j$, i.e. the set of all players in the dataset $T$.

As other approaches have done in the past, Elo++ uses a global parameter $\gamma$ that captures the advantage of the white player. The parameter $\gamma$ is added to the rating of the player who is playing white. The formula for predicting the outcome of a game between a white player $i$ and a black player $j$ is given by the following logistic curve:

$$\hat{o}_{ij} = \frac{1}{1 + e^{r_j - (r_i + \gamma)}}.$$

Such logistic curves are in the heart of many Elo-like rating systems. The intuition is that if the rating of $i$ is much bigger than the rating of $j$, then the predicted outcome $\hat{o}_{ij}$ goes to 1 (i.e. white wins). When the ratings are close then the predicted outcome goes to 1/2 (i.e a draw) and when $i$ is much worse than $j$, then it goes to 0 (i.e. black wins).

The performance of predicted outcomes in the kaggle competition was evaluated using a *variant* of the Root Mean Square Error (RMSE):

$$\text{RMSE} = \sqrt{\frac{1}{|S|} \sum_{ij \in S} (\hat{o}_{ij} - o_{ij})^2}$$

where $S$ is the private hold-out dataset. The RMSE variant used in the kaggle competition is called Player/Month-aggregated RMSE (PM-RMSE). Without going into details, the main difference is that PM-RMSE aggregates the prediction errors $\hat{o}_{ij} - o_{ij}$ by month and player, ignoring the color of the player.

## 3 Elo++ Details

In the following, we discuss in detail how Elo++ computes the ratings $r_i$ of the players, using the basic notation introduced in Section 2.

### 3.1 Time Scaling

Elo++ takes into account the time of each game. The reason is that old games have less importance than newer games. For example, young players tend to improve fast, while top players maintain a stable high rating over their career before they get older and start progressively to lose some of their competitive strength. Capturing these temporal dynamics is important for any rating system; in Elo++ a simple weighting scheme over the recency of the games has been employed.

Let's assume that $t_{min}$ is the time the earliest game happened in our dataset (i.e. month 1 in the training dataset) and $t_{max}$ is the time the latest game took place (i.e. month 100). For a game that happened at time $t_{ij}$, the weight that worked best in the Elo++ experiments is:

$$w_{ij} = (\frac{1 + t_{ij} - t_{min}}{1 + t_{max} - t_{min}})^2.$$

During the training of the ratings, the importance of each game is scaled with the corresponding $w_{ij}$. The details are discussed in Section 3.3. The idea is inspired by Chessmetrics [2], which uses a similar approach. The scaling worked fine for Elo++, without having to throw-away any games (many rating systems discard old games as irrelevant).

### 3.2 Neighbors

Each rating $r_i$ for a player $i$ is trained based on a relatively small number of games for player $i$. Therefore, the potential for overfitting the ratings over the training dataset is large. To avoid this, Elo++makes the assumption that most games between chess players happen in tournaments, where players of comparable strength play against each other. It is rare, that a weak player participates in a tournament of strong players (and vice-versa). This observation implies that the rating of a player and the ratings of his opponents are strongly correlated. In this section, we define a weighted average of the opponents of a player $i$. In Section 3.3 we show how to use regularization to "pull" each rating $r_i$ close to its weighted average.

Let's define the *neighborhood* $N_i$ of a player $i$ as the multiset of all opponents player $i$ has played against, regardless of color. It's defined as a multiset, since a player may have played the same opponent many times or with different colors. As we described, we expect that there is a strong connection between the average rating of $N_i$ and the rating of player $i$. One affects the other in a chicken-and-egg way. In addition, the more games a player has played, the more confident we are about his rating. Similarly, the more recent the games are, the more accurate his rating is. As discussed in Section 3.1, each game is associated with a weight $w_{ij}$, that depends on the time the game took place. For every player $i$, with a neighborhood $N_i$ we define the following weighted average:

$$a_i = \frac{\sum_{k \in N_i} w_{ik} r_k}{\sum_{k \in N_i} w_{ik}},$$

where the sum runs over all the neighbors $k$ of player $i$, regardless of the colors of $i$ and $k$. Intuitively, $a_i$ is close to the average rating of the opponents that have played many recent games against $i$. The ratings of old and infrequent opponents are scaled down using the weights $w_{ik}$. In Elo++, this weighted average $a_i$ is used as a neutral prior for regularization (see Section 3.3) and it is the major component, that separated Elo++ from other techniques in the competition.

## 3.3 Rating Update Formulas

In this section, we discuss exactly how Elo++ computes the ratings $r_i$. The logistic curve, that Elo++ uses for predicting the outcome of a game between white player $i$ against black player $j$, is:

$$\hat{o}_{ij} = \frac{1}{1 + e^{r_j - (r_i + \gamma)}}$$

The *total loss* $l$ between the predicted and the actual outcomes in Elo++ is defined as:

$$l = \sum_{ij \in T} w_{ij} (\hat{o}_{ij} - o_{ij})^2 + \lambda \sum_{i \in D} (r_i - a_i)^2.$$

The total loss takes into account the recency of the games, the differences between outcomes and predictions and finally the neighbors' ratings. The recency is expressed through the $w_{ij}$ weights, and the differences between outcomes and predictions are expressed through the $(o_{ij} - \hat{o}_{ij})^2$ summation. The $l_2$ regularization "penalizes" ratings $r_i$ that are disproportionally away from the corresponding neighbors average $a_i$.

Training in Elo++ means optimizing the ratings so that the total loss is minimized. Intuitively, the optimal ratings will (a) minimize the differences between recent predicted and actual outcomes, and (b) be as close to their corresponding neighbor averages $a_i$ as possible. The regularization introduces an interesting chicken-and-egg relationship between $r_i$'s and $a_i$'s. Note, that (a) $a_i$'s are defined in terms of weighted $r_k$'s in the neighborhood of $i$ and (b) the optimal ratings $r_i$'s are "not allowed" to diverge a lot from their corresponding $a_i$'s, unless there is sufficient evidence in the training data. The "strength" of this evidence is captured by the global parameter $\lambda$. The discussion around the nature and optimal value for $\lambda$ is in Section 3.4.

The optimization problem defined by the total loss is non-linear and non-convex. For example, if $\lambda$ is set to zero and the neighborhood is not taken into account, then it is easy to see that there is an infinite number of optimal minima; let's assume that there is an optimal minimum, one can construct infinitely many equivalent optimal minima, just by adding any constant to the optimal ratings. The regularization and the neighborhood complicate further the surface of the optimization problem, by introducing many local minima. To address this optimization, Elo++ uses a stochastic gradient descent technique [10].

**Algorithm 1** Elo++ Ratings Computation

**Require:** $T$: training dataset, $\gamma$: white's advantage, $\lambda$: regularization constant {Section 3.4}
**Require:** $P$: total number of iterations
  for all players $i$, $r_i \leftarrow 0$
  for all games in $T$ compute time weights $w_{ij}$ {Section 3.1}
  **for** $p = 1$ **to** $P$ **do**
    for all players $i$, compute neighbor averages $a_i$ {Section 3.2, $|N_i|$ is the neighborhood size}
    $\eta \leftarrow ((1 + 0.1P)/(p + 0.1P))^{0.602}$ {learning rate}
    **for all** *shuffled* tuples $\langle i, j, t_{ij}, o_{ij} \rangle$ in $T$ **do**
      $\hat{o}_{ij} = 1/(1 + exp(r_j - (r_i + \gamma)))$
      $r_i \leftarrow r_i - \eta[w_{ij}(\hat{o}_{ij} - o_{ij})\hat{o}_{ij}(1 - \hat{o}_{ij}) + \frac{\lambda}{|N_i|}(r_i - a_i)]$
      $r_j \leftarrow r_j - \eta[-w_{ij}(\hat{o}_{ij} - o_{ij})\hat{o}_{ij}(1 - \hat{o}_{ij}) + \frac{\lambda}{|N_j|}(r_j - a_j)]$
    **end for**
  **end for**
  **return** all ratings $r_i$

Intuitively, the main idea is to repeatedly update the ratings using noisy estimates of the gradient of the total loss. Such gradient estimates are computed efficiently using a single random tuple at a time. Although the rating updates are noisy, one can show that —under certain regularity conditions— the noise cancels out, and the stochastic process converges to a local minimum.

The Elo++ algorithm for computing the ratings is summarized in Algorithm 1. First Elo++ sets all the ratings to zero (i.e. $r_i \leftarrow 0$), and then performs repeated iterations over the training dataset. Before each iteration, all the training tuples are shuffled (i.e their order is randomized) and the average $a_i$'s are computed. The iteration consists of updating both $r_i$ (for the white player) and $r_j$ (for the black player) after each tuple $\langle i, j, t_{ij}, o_{ij} \rangle$ using the following *rating update formulas*:

$$r_i \leftarrow r_i - \eta[w_{ij}(\hat{o}_{ij} - o_{ij})\hat{o}_{ij}(1 - \hat{o}_{ij}) + \frac{\lambda}{|N_i|}(r_i - a_i)]$$

$$r_j \leftarrow r_j - \eta[-w_{ij}(\hat{o}_{ij} - o_{ij})\hat{o}_{ij}(1 - \hat{o}_{ij}) + \frac{\lambda}{|N_j|}(r_j - a_j)]$$

where $\eta$ is the learning rate, and $|N_i|$, $|N_j|$ are the sizes of the neighborhoods of players $i$ and $j$ respectively. To simplify the update formulas, we made the assumption that all the averages $a_i$, $a_j$ are kept constant through the iteration (they are computed before each iteration and do not change during the iteration). Otherwise, every time a rating $r_i$ was updated, all the affected $a_k$ in the neighborhood $N_i$ would have to be updated as well; this is computationally expensive and complicates unnecessarily the update formulas.

The learning rate $\eta$ (see [10] for details) in Elo++ is defined as:

$$\eta = (\frac{1 + 0.1P}{p + 0.1P})^{0.602},$$

where $P$ is the maximum number of iterations, and $p$ is the current iteration number.

In most of the experiments, the described stochastic gradient descent got to the vicinity of the minimum after five iterations, and converged after $P = 50$ iterations. Simpler formulas for the learning rate (like $\eta = 1/p$) also converge, but slower than the one described above. Other alternatives for this optimization problem would be to use a deterministic optimization library, like for example L-BFGS-B [8]. However, such deterministic optimizations (a) converge slower and (b) tend to overfit more than the stochastic approach described here.

Elo++ also employed an early-out approach; i.e while training we keep track of the corresponding cross-validation loss (the loss on a cross-validation dataset that doesn't participate in the training). When the cross-validation loss starts increasing, then there is indication that overfitting is happening. For the dataset of the competition, the early-out approach provided minor benefits, probably because the $l_2$ regularization technique already reduced overfitting significantly.

The whole script in R [9] required only around 100 lines of code, demonstrating the power of R in relatively complex analytical tasks. Overall the PM-RMSE obtained using Elo++ on the private hold-out dataset was 0.69356[1], outperforming in the competition all approaches including TrueSkill, Chessmetrics, Glicko, Elo and PCA.

### 3.4 Global Elo++Parameters

In Section 3.3 we showed how the computation of the optimal ratings $r_i$ works, given two global parameters: $\gamma$ (white's advantage) and $\lambda$ (regularization). In this section, we discuss these global parameters and their optimal values.

The global parameters that worked best in the experiments are: $\gamma = 0.2$ and $\lambda = 0.77$. It is quite interesting that the optimal $\lambda$ value is so large; it means that there is a very strong correlation between the ratings of the neighbors and the actual rating of a player in the kaggle dataset.

Both parameters $\lambda$ and $\gamma$ were optimized through cross-validation. For different chess datasets their optimal value should be close to the numbers we depict here; unless these datasets exhibit completely —and surprisingly— different characteristics.

Parameter $\lambda$ *could* be sensitive to the size $|T|$ of the training dataset and the cardinality $|D|$ of the domain of players. The author suspects that if their relative sizes change, then the optimal $\lambda$ might change as well. If, for example, there are many more games per distinct player (i.e. $T$ grows disproportionally faster than $D$) then the optimal parameter $\lambda$ could get smaller. How exactly the optimal $\lambda$ is affected by $|D|$ and $|T|$, is an open question that deserves a thorough study.

## 4 Quality Comparison

In this section, we discuss the quality of the actual ratings returned by Elo++. After the end of the competition, a list with the actual names and Elo ratings of all players in the competition datasets was made available. In the following, we'll refer to this list as the *Elo list*. We compare and discuss the ratings returned by Elo++ against these widely used Elo ratings.

First we "normalize" the ratings returned by Elo++ to bring them in the same order of magnitude as the Elo ratings. The normalization helps compare directly the ratings. It is required because (a) Elo uses a base-10 logistic curve instead of Elo++'s base-$e$, (b) Elo scales the ratings with a factor of 400 and (c) Elo has constraints on the ratings of average-level and master-level players. The normalization simplifies to multiplying each Elo++ rating with a constant factor of $400 \log_{10} e \approx 173.72$, and adding the constant 2338 (so that both distributions have the same mean). An interesting point to make is that the scaled white's advantage is $400\gamma \log_{10} e \approx 34.7$. In other words just by playing white, one has an average advantage of 34.7 rating points.

In Figure 1, we show the histograms of the Elo and Elo++ ratings for all players in the dataset. The x-axis depicts respectively the Elo and Elo++ ratings, while the y-axis depicts the number of distinct players with a rating in the corresponding bucket of the x-axis. We observe that Elo++ ratings are (a) more symmetrical and (b) more concentrated around the mean rating. This is due to the Elo++ regularization; it avoids "micro"-adjusting the ratings, unless there is enough statistical evidence in the data.

In Figure 2, we draw the scatter plot of the Elo++ ratings over the Elo ratings. Each point in the scatter plot corresponds to a distinct player, whose Elo rating is on the x-axis, and Elo++ rating is on the y-axis. We observe that there is a very strong correlation for high ratings. In other words, players with high Elo ratings have also high Elo++ ratings and vice-versa. We emphasize though, that although the top-Elo list and the top-Elo++ list contain roughly the same top players, their relative order is different. For example, Anand Viswanathan holds the highest Elo rating in the Elo list, with an Elo rating of 2801, but is second in the Elo++ list with an Elo++ rating of 2802. The author doesn't want to draw more conclusions from the Elo list, since it probably corresponds to more games and players, than the dataset that Elo++ used. However, the comparison helps to make the point that Elo++ provides comparable to Elo ratings and could be used to enhance existing Elo lists.

---

[1] This is better than the 0.69477 reported on the kaggle website, because of a bug in the script that was fixed after the competition was finished, while the author was double-checking and repeating all the experiments.

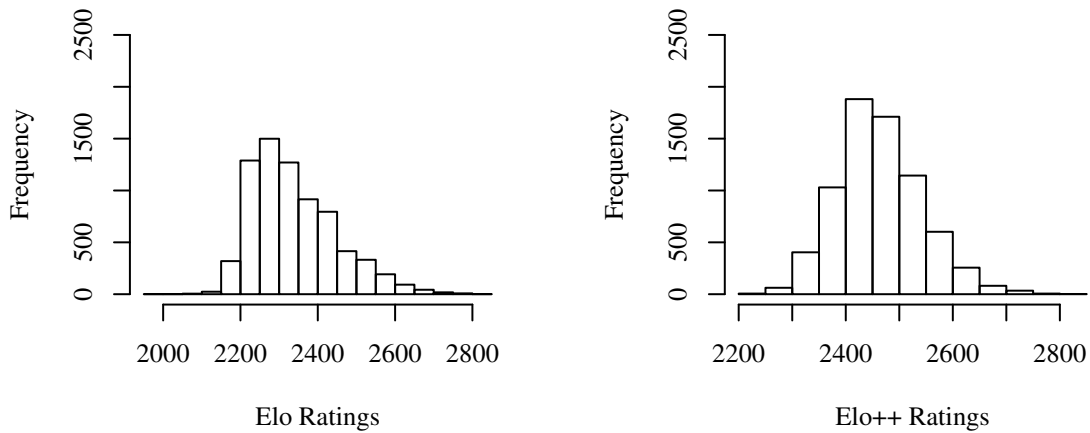

Figure 1: Histogram Comparison

The correlation is much weaker for medium and low ratings. Since Elo and Elo++ are derived in different ways, we expect such differences. We point out that in certain cases the ratings for the same player, differ by more than 400 rating points. The exact reason behind such large rating differences deserves a thorough study, beyond the scope of this article.

## 5 Conclusions

My intent in writing this article is to celebrate the conclusion of the first kaggle competition around chess ratings. This was the most popular kaggle competition to date in terms of participation. Hopefully, a followup competition will allow further improvements in this interesting area.

The science of chess rating systems is the prime beneficiary of the competition. Many new people (including the author) became involved in the field and made their contributions. Out of the numerous new algorithmic contributions that are discussed in the forums, I would like to highlight (a) the importance of properly regularizing the ratings and (b) the basic logistic curve, that still holds strong. Although more sophisticated algorithmic aspects and models are possible, an accurate treatment of the basics is at least as significant as coming up with new modeling breakthroughs.

The winning submission happened less than four weeks after the competition had started. However, the discrepancies between my own cross-validations and the leader-board made me believe that much more complicated models were required, and I soon abandoned the basic principles of the winning submission. I tried more complicated models, while relaxing the regularization efforts. In retrospect, this was a bad decision on my part; I should have realized earlier the importance of regularization, and the potential of overfitting for such a small test data set.

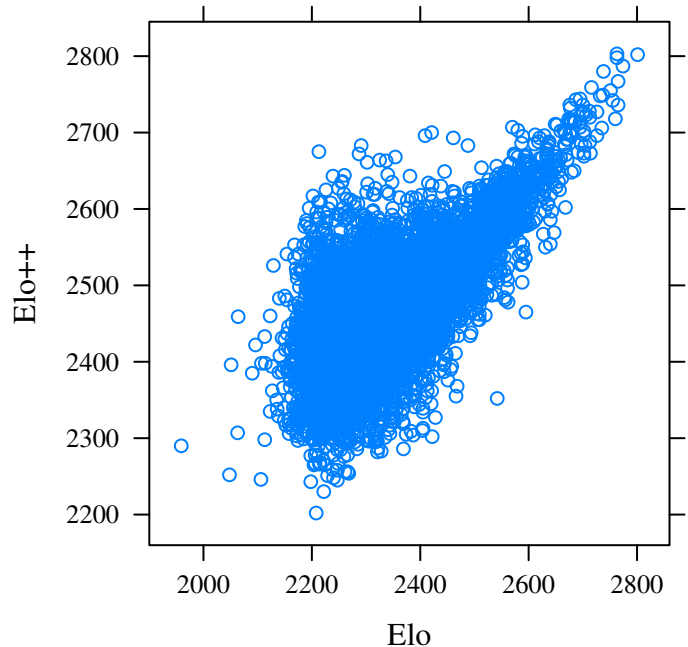

Figure 2: Scatter Plot of Elo++ w.r.t. Elo

The author was lucky to win the competition. First, I'd like to thank everyone involved in setting up this excellent competition. Second, all the competitors deserve congratulations for their contributions. I would like to thank especially those who published their results, participated in the forums and provided their intuitions.